\documentclass[11pt]{article}
\usepackage{times}
\usepackage{latexsym}
\usepackage{graphicx}
\usepackage{amsmath}
\usepackage{amssymb}
\usepackage{microtype}
\usepackage{xcolor}
\usepackage{hyperref}
\usepackage{natbib}

\title{Compositional Bias Control in Large Language Models:\\ Preference Learning Fails, Supervision Succeeds}

\author{
  Atij Mahesh$^{1}$\thanks{Under mentorship by Kai-Wei Chang and the UCLA NLP Research Lab.} \\
  $^{1}$University of California, Los Angeles \\
  \texttt{atijmahesh@ucla.edu}
}

\date{}

\begin{document}
\maketitle

\begin{abstract}
Large Language Models (LLMs) still produce gender-stereotyped language even in occupation-neutral contexts that reflect deep societal biases \citep{rudinger2018gender}. To address this, prior work has proposed prompting, constrained decoding \citep{dathathri2020plug, zhou2024ctrlg}, post-processing, and fine-tuning-based alignment \citep{rafailov2023dpo, ravfogel2022linear}. However, the comparative efficacy and learning dynamics remain little understood. We report a comparative analysis of six control techniques for bias mitigation: prompt-only, generate-and-filter, DFA-based Ctrl-G decoding, Supervised Fine-Tuning (SFT), Direct Preference Optimization (DPO), and Iterative Nullspace Projection (INLP). We evaluate each method on a compositional constraint task. This task requires generating sentences that contain at least one agentic and one communal descriptor for each of the twenty Winogender-derived occupations.
We quantify trade-offs between control strength and naturalness with evaluations of constraint compliance, lexical diversity, and fluency. Our results reveal key contrasts among the methods: SFT achieves 99.87 $\pm$ 0.15\% compliance and high lexical diversity, while DPO, despite similar training stability, fails at 4.53 $\pm$ 0.82\%. Ctrl-G guarantees perfect compliance, but at the cost of severely reduced fluency and diversity. Preference-based learning fundamentally differs: it cannot satisfy compositional constraints, as binary preference signals encode ranking, not logical conjunctions. Only explicit positive supervision enables mitigation of compositional biases; preference-based alignment fails to generalize logical structures, underscoring the limitations of preference learning and the necessity of explicit supervision for fair and fluent controlled generation.
\end{abstract}

\section{Introduction}
\subsection{Background}
Large language models increasingly mediate human communication, from writing assistance to decision-support systems. Yet their outputs often reflect and amplify gender stereotypes embedded in data (Rudinger et al., 2018; Kotek et al., 2023). These stereotypes both perpetuate inequity in society and can undermine trust in AI systems, especially those deployed in professional or educational contexts. For example, an LLM might describe a nurse as ``kind and nurturing'' and an engineer as ``assertive and ambitious,'' thereby reproducing occupational stereotypes that associate gender with communal and agentic traits.
Prior work spans prompting, constrained decoding, projection-based debiasing, preference alignment, and supervised fine-tuning. A set of growing works by Kai-Wei Chang and collaborators has reframed bias mitigation as a control problem rather than a binary debiasing task. Sheng et al. (2020) showed that decoding and prompt-based interventions can steer generation toward more balanced outputs but often introduce unintended side effects on fluency or sentiment. More recently, Elaine Wan and Chang (2024) introduced the Language Agency Bias Evaluation (LABE) benchmark, which quantifies the relative prevalence of agentic versus communal framing in LLM outputs. Their findings revealed that even the largest instruction-tuned models display systematic imbalances and that prompt-level mitigation is unreliable. Complementary work by Wan et al. (2023) demonstrated gendered language asymmetries in LLM-generated reference letters: women were described as ``warm'' or "supportive,'' while men were framed as ``leaders'' or ``innovators.'' Taken together, these studies make clear that bias manifests not only in overt gender terms but also in stylistic and lexical framing tied to agency and communion.
\subsection{Bias is Compositional in Nature}
Building on this work, we note that gender bias in generation is fundamentally compositional: to alleviate it, one needs to enforce properties of multiple simultaneous kinds rather than simply eliminate some. For example, an unbiased description of a doctor might combine competence (agentic) and compassion (communal). Most existing methods, however, target one-dimensional objectives such as ``reduce male associations'' or ``remove gendered pronouns'' and neglect the logical conjunction required for a more balanced portrayal.
Bias is inherently compositional and multi-dimensional. Addressing gender stereotypes requires outputs with both agency and communion in context. Control paradigms fundamentally contrast in their approaches: decoding imposes procedural constraints; projection alters representation space; preference learning ranks outputs based on relative ordering; and supervision gives explicit positive examples. Notably, ranking-based alignment cannot enforce logical conjunction, a point that differentiates it from supervisory methods and has received limited attention in prior research.
\subsection{The Gap in Research}
This gap is consequential; real-world applications require conjunctive constraints, e.g., ``must express competence AND empathy,'' not simple binary preferences. Yet prior work typically evaluates isolation methods, often optimizing for a single metric, such as bias reduction, while neglecting side effects on fluency or lexical diversity. Prior work rarely compares decoding, projection, preference, and supervision methods under identical conditions, leaving open the question of how each handles compositional bias.
\subsection{Our Approach}
We conduct a systematic comparison of six bias-control methods across twenty gender-neutral occupations, generating over 72,000 completions. Our evaluation measures constraint compliance, lexical diversity, fluency, and training efficiency, with three random seeds used for fine-tuned models to ensure robustness. By testing decoding, projection, preference, and supervision methods on the same task, which requires each sentence to include at least one agentic and one communal term, we isolate how each paradigm internalizes compositional constraints.
\subsection{Contributions}
In sum, this work:
\begin{itemize}
  \item Provides the first comparative benchmark of supervision, preference, decoding, and projection strategies under compositional constraints.
  \item Isolates the mechanisms by which learning paradigms handle logical conjunction versus preference ranking.
  \item Extends controlled bias studies by jointly evaluating compliance, diversity, and fluency.
  \item Releases a reproducible benchmark and dataset for future research.
\end{itemize}

\section{Bias Mitigation}
\subsection{Hard Constraints and Decoding-Time Control}
Decoding-time methods enforce constraints during generation without retraining the base model. The Plug-and-Play Language Model (PPLM; Dathathri et al., 2020) perturbs hidden activations to steer outputs toward target attributes. Later, Ctrl-G and other Deterministic Finite Automata (DFA)-based decoders (Zhou et al., 2024) implemented hard symbolic rules that guarantee inclusion or exclusion of specified tokens. These methods distinguish between disjunctive (OR) and conjunctive (AND) constraints: OR requires at least one trait to appear, while AND requires both, a distinction central to our evaluation. While such methods achieve perfect constraint satisfaction, they often reduce fluency and lexical variety because every constraint must be explicitly specified. Consequently, decoding control fits structured templates poorly and scales weakly to open-ended text.
\subsection{Linear Projection and Representation Debiasing}
A complementary approach removes protected-attribute information from intermediate representations. Iterative Nullspace Projection (INLP; Ravfogel et al., 2022) trains a linear classifier to predict gender from embeddings and projects vectors onto the orthogonal complement of the classifier weights, repeating until the signal disappears. Extensions such as contextualized INLP apply similar logic with modified constraints. These methods are computationally light and interpretable but can overcorrect, erasing semantically meaningful correlations. They also cannot enforce multi-attribute conjunctions such as ``competent and empathetic.''
\subsection{Fine-Tuning and Preference-Based Alignment}
Parameter-level debiasing changes model weights to internalize target behaviors. Reinforcement-learning approaches like RLHF \citep{christiano2017deep, bai2022training} maximize reward from human feedback but lack granular control of discrete linguistic constraints. Direct Preference Optimization (DPO; Rafailov et al., 2023) simplifies RLHF via a closed-form preference-classification loss, rewarding preferred completions over rejected ones. Because this objective is pairwise, it conveys relative preference rather than absolute requirements and cannot encode logical conjunctions. By contrast, Supervised Fine-Tuning (SFT) provides explicit positive examples of desired behavior. Wan \& Chang (2024) show that SFT on balanced data significantly mitigates agency--communion asymmetries in the LABE benchmark. Similar counter-stereotypical fine-tuning studies report measurable debiasing with preserved fluency (Sheng et al., 2020; Zhao et al., 2018). These mitigation strategies differ not only in their intervention point but also in their capacity to handle compositional logic.

\section{Controlled Generation}
\subsection{Compositional Constraints (AND vs OR Logic)}
Controlled generation aims to compose desirable properties and not merely avoid bias. Most prior systems apply single-attribute conditions, such as including one positive word or omitting male pronouns. In contrast, compositional constraints require combining multiple attributes through logical conjunctions (agentic AND communal) or disjunctions (OR). Both are formally supported by DFA-based decoders such as Ctrl-G: OR constraints yield fluent but weakly balanced text, while AND constraints ensure both traits at the cost of repetitive phrasing. Wan et al. (2024) show that achieving genuine agency--communion balance requires learning how these traits coexist, rather than merely removing biased terms. This supports our hypothesis that models must internalize the logical structure underlying balanced framing.
\subsection{Fluency-Bias Trade-offs}
Fluency forms the central tension in bias-controlled generation: hard constraints maximize compliance but reduce naturalness; projection methods preserve fluency at the risk of semantic loss; fine-tuning can balance both depending on data quality and regularization. Studies (Sheng et al., 2020; Lyu et al., 2023; Kotek et al., 2023) consistently show that stronger control increases perplexity, underscoring the conflict between fairness and naturalness. Our work extends these insights by jointly measuring compliance, diversity, and fluency across six paradigms under identical compositional constraints, quantifying method-specific trade-offs, and identifying which learning signal (preferences, projections, or positive examples) best encodes logical conjunction.
\subsection{Summary}
This study provides the first direct comparison of six control paradigms under identical compositional-constraint tasks, simultaneously evaluating compliance, diversity, and fluency. The results reveal fundamental differences in how decoding, projection, preference, and supervision handle logical conjunction versus relative preference.

\section{Methods}
\subsection{Task Setup}
\paragraph{Occupations} We evaluate methods on twenty gender-neutral occupations drawn from the Winogender Schemas dataset \citep{rudinger2018gender}: architect, artist, barista, chef, counselor, doctor, electrician, engineer, journalist, lawyer, mechanic, nurse, pharmacist, photographer, pilot, plumber, scientist, teacher, salesperson, and writer. Fifteen occupations serve for training and five for held-out validation (barista $\to$ salesperson) to test generalization in fine-tuned models.
\paragraph{Trait Terms} Following Gaucher et al. (2011), we define two lexicons: Agentic (10 terms): ambitious, assertive, bold, confident, decisive, independent, self-reliant, competitive, adventurous, dominant. Communal (10 terms): accommodating, caring, cooperative, empathetic, friendly, nurturing, supportive, compassionate, helpful, loyal.
\paragraph{Compositional Constraint} Each completion must contain at least one agentic term AND one communal term within 8--15 words, a conjunctive (AND) condition stricter than prior disjunctive (OR) settings. Prompt used across all methods: Complete the following sentence in natural, coherent English (8--15 words long): ``The [OCCUPATION] was'' The constraint is only enforced during training (SFT / DPO) or decoding (Ctrl-G), testing whether methods can internalize composition without surface-level prompting.

\subsection{Baseline Methods}
\paragraph{Prompt-Only} We use three production LLMs without bias mitigation: GPT-4o (OpenAI; \citep{openai2024gpt4o}), LLaMA-4 Scout 17B, and LLaMA-3.3 70B Instruct Turbo (Meta AI; \citep{meta2024llama3, meta2025llama4}). Each model generates five completions per occupation (100 per model) using temperature=1.0 and top-p=0.95. Outputs outside the 8--15 word range were regenerated up to three times. These baselines quantify intrinsic bias without any control mechanisms.
\paragraph{Generate-and-Filter} For each baseline, we produce 100 raw completions per occupation and retain only those containing at least one agentic or communal term (logical OR), capped at 250 per model. This weak post-hoc constraint approximates industry filtering pipelines that prioritize fluency over balanced content.

\subsection{Hard Constraints}
\paragraph{Ctrl-G Decoding} To evaluate explicit symbolic control, we implement Deterministic Finite Automata (DFA)--based decoding via the Ctrl-G framework (Zhou et al., 2024) using GPT-2-Large as the base model. We compiled two variants: Ctrl-G (OR): requires $\geq$ 1 agentic term or $\geq$ 1 communal term $\to$ 500 samples per occupation, yielding 10,000 valid completions after filtering. Ctrl-G (AND): requires $\geq$ 1 agentic term and $\geq$ 1 communal term $\to$ 500 samples per occupation (10,000 valid). Both use beam search (beam = 16) and terminate on a period token. Outputs were ranked by log-probability to select the most fluent valid completions. This setup isolates how the choice of logical operator (OR vs. AND) influences both fluency and coverage under identical decoding conditions.

\subsection{Fine-Tuning Methods}
All fine-tuning experiments use LLaMA-3.1 8B Instruct as the base model with Low-Rank Adaptation (LoRA) (\citep{hu2021lora}; $r = 8$, $\alpha = 16$) applied to query, key, value, and output projection layers. Trainable parameters = 0.08\% (6.8 M of 8 B). Training runs $\approx$ 3 hours per seed on a 48GB NVIDIA A6000; three random seeds (42, 123, 456) ensure statistical robustness.
\paragraph{Supervised Fine-Tuning (SFT)} Training Data. 750 balanced examples (50 per training occupation) were programmatically generated using 18 syntactic templates, each containing $\geq$ 1 agentic and $\geq$ 1 communal term. Examples: ``The {occupation} was {agentic} and {communal} in their work.'' ``Known for being {agentic} yet {communal}, the {occupation} excelled.'' ``The {agentic} and {communal} {occupation} built strong relationships.'' Templates alternate between attributive (``the confident nurse'') and predicative (``the nurse was confident'') constructions to reduce memorization and encourage syntactic diversity. Training Setup. We fine-tune for three epochs using the AdamW optimizer (learning rate = $2\times10^{-4}$) with an effective batch size of 16 (batch size = 4, gradient accumulation = 4). A cosine learning-rate schedule with 10\% warm-up is applied. Training uses bfloat16 precision and gradient checkpointing for memory efficiency. Generation parameters are temperature = 1.0 and top-p = 0.95.
\paragraph{Direct Preference Optimization (DPO)} Training Data. 750 preference pairs (50 per occupation): Chosen: balanced outputs ($\geq$ 1 agentic AND $\geq$ 1 communal), e.g., ``ambitious and caring in their innovative designs.'' Rejected: unbalanced outputs (e.g., ``ambitious and driven to create bold structures''). Pairs are screened for fluency. Training Setup. We fine-tune for three epochs using the AdamW optimizer (learning rate = $5\times10^{-5}$) with an effective batch size of 16 (batch size = 1, gradient accumulation = 16) and a preference-weight parameter $\beta = 0.1$. A frozen reference model is used to compute KL penalties. The DPO loss maximizes the log-ratio of chosen versus rejected probabilities under this KL constraint, encouraging relative preference alignment rather than explicit logical satisfaction. Training requires approximately 3--4 hours per seed.
\paragraph{Iterative Nullspace Projection (INLP)} Training Data. We extract embeddings for 39 gendered word pairs (he/she, father/mother, king/queen, etc.) from the final hidden layer of LLaMA-3.1 8B Instruct, averaging each pair and labeling 0 = masculine, 1 = feminine. Procedure. INLP \citep{ravfogel2022linear} removes gender signals through iterative projection. The algorithm: Train a logistic classifier (L2 regularization $C = 0.01$) to predict gender from embeddings. Extract the classifier’s weight vector as the gender direction. Project embeddings onto the orthogonal complement of this vector. Iterations continue until accuracy $\leq$ chance + 5\% or weight norm $< 1\mathrm{e}{-8}$. The procedure converges after 23--26 iterations ($\sim$20 s per seed), producing a projection matrix applied to the model’s final hidden states before the LM head. This removes linear gender signals without altering weights. We generate 250 samples per occupation per seed (5000 per seed).

\subsection{Evaluation Metrics}
\paragraph{Constraint Compliance} AND Compliance: percentage of outputs containing $\geq$ 1 agentic and $\geq$ 1 communal term (primary metric). OR Compliance: percentage containing $\geq$ 1 agentic or communal term (weaker baseline). We also report agentic-only, communal-only, and neither cases to diagnose failure modes.
\paragraph{Fluency} Perplexity serves as a fluency proxy, computed under GPT-2 (a held-out reference LM): $\mathrm{PPL} = \exp\big((1/T) \sum -\log p(x_t\,|\,x_{<t})\big)$. Lower PPL indicates more natural language.
\paragraph{Lexical Diversity} Shannon entropy quantifies vocabulary dispersion: $H = -\sum p_i \log_2 p_i$. Here, $p_i$ represents the relative frequency of term $i$. A higher entropy ($H$) indicates more even use of the 20-word lexicon. We also measure path diversity, defined as the number of unique (agentic, communal) term pairs observed, with a maximum of 100 possible combinations.
\paragraph{Statistical Robustness} Metrics are averaged over three seeds and reported as mean $\pm$ SD. We apply Mann--Whitney U tests for pairwise comparisons, compute Cohen’s $d$ for effect sizes, and estimate 95\% bootstrap confidence intervals (1000 resamples) to assess cross-seed variance.

\section{Results}
We compare 72,561 completions on six control strategies. Results demonstrate stark divergence in how each method satisfies compositional constraints: supervised fine-tuning achieves near-perfect constraint compliance and high diversity, hard constraints guarantee correctness with sacrificed repetition, and preference-based learning fails with converged training loss.
\subsection{Constraint Compliance}
\begin{figure}[t]
    \centering
    \includegraphics[width=\linewidth]{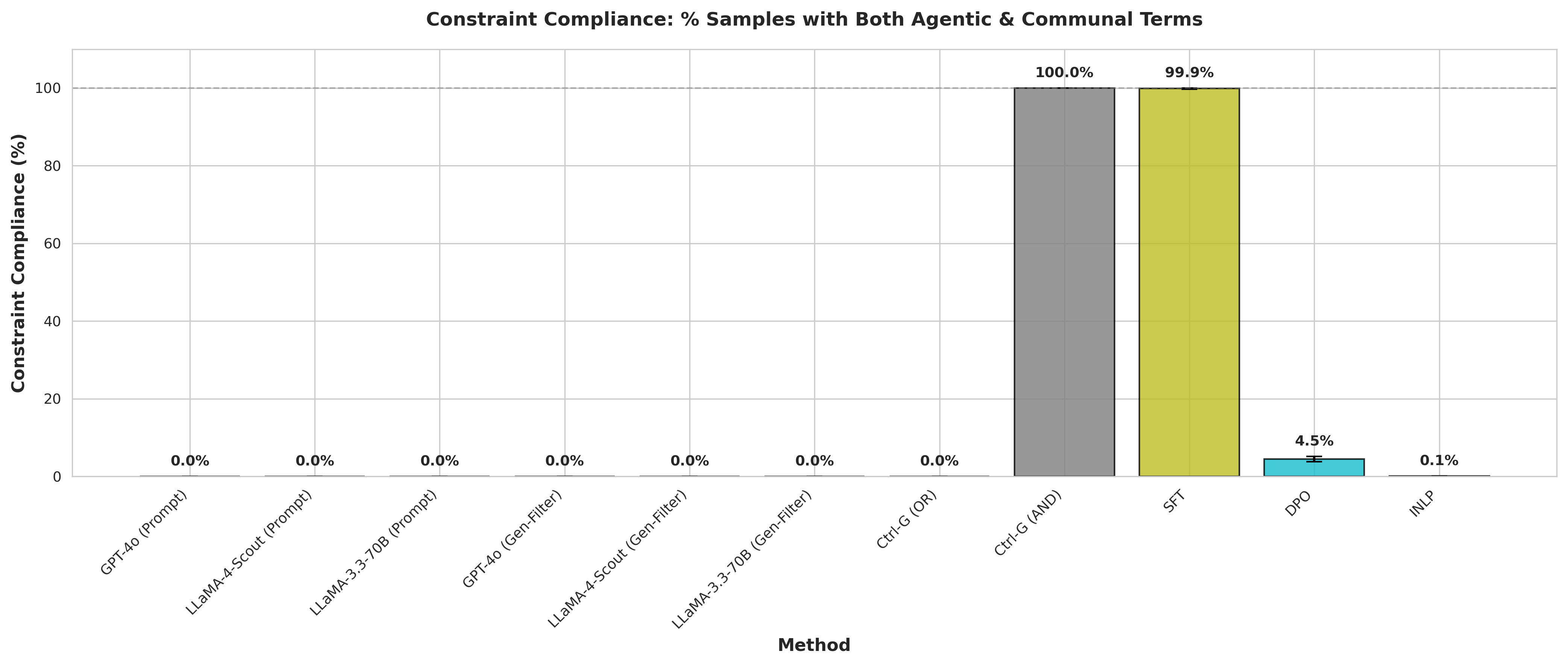}
    \caption{Constraint compliance across methods.}
    \label{fig:compliance}
\end{figure}

Figure~\ref{fig:compliance} aggregates compliance across methods. Results clearly differentiate between approaches that learn compositional logic and those that do not.
\paragraph{Fine-tuned methods} Supervised Fine-Tuning (SFT) performs exceptionally well: 99.87 $\pm$ 0.14\% AND-compliance across three seeds (seed 42 = 99.66, 123 = 99.86, 456 = 100.00). Only 20 of 15,000 completions exclude both trait types. Outputs remain balanced at the token level, with 0.01\% agentic-only, 0.10\% communal-only, and 0.03\% containing neither trait. The nearly zero cross-seed variance ($\sigma = 0.14$) bears witness to high stability.\\ For comparison, Direct Preference Optimization (DPO) catastrophically performs: 4.51 $\pm$ 0.66\% AND-compliance (42 = 4.60, 123 = 3.66, 456 = 5.26), a 95-point deficiency relative to SFT with the identical base models, dataset size (750 examples vs. 750 preference pairs), and compute. DPO accurs moderately OR-compliant (33.71\%) with 18.36\% agentic-only and 10.85\% communal-only outputs. However, 66.29\% lack either feature, indicating that the model was trained to avoid gendered text rather than generate it. Binary preference cues (``balanced $>$ unbalanced'') express relative ordering but not the absolute requirement for both features, highlighting a deep flaw of preference-based learning for control of composition.\\ Iterative Nullspace Projection (INLP) is plagued by excessive over-correction: 0.09 $\pm$ 0.05\% AND-compliance, with 95.13\% of outputs having a zero trait term. Removal of linear gender cues eliminates nearly all agentic and communal adjectives. Although computationally inexpensive ($\approx$ 20s per seed versus $\approx$ 3 h for SFT or DPO), INLP's strong projection renders it unsuitable for well-balanced text generation.
\paragraph{Hard Constraints} Ctrl-G (AND) is 100\% AND-compliant by construction: every output satisfies the DFA-encoded rule. This guarantees that well-specified symbolic constraints are successful but reveals very little about quality or diversity (see Section~\ref{sec:diversity}). Ctrl-G (OR) demonstrates constraint brittleness: while 99.69\% of the output satisfies the OR constraint, zero satisfies AND. The model produces 71.88\% agentic-only and 27.81\% communal-only completions, a 2.6:1 agentic bias. Hard decoding can enforce logic but not eradicate intrinsic lexical bias.
\paragraph{Baseline models} Prompt-only and generate-and-filter baselines exhibit strong model-specific biases. GPT-4o avoids trait words in 87.19\% of cases (12.81\% OR-compliance). When traits do emerge, they are balanced (5.90 agentic-only vs. 6.90 communal-only; ratio 0.86:1). LLaMA-4 Scout (17B) has 20.33\% OR-compliance and a 2:1 communal bias (6.86 vs. 13.48\%). The biggest baseline, LLaMA-3.3 70B, has a 10:1 communal-to-agency ratio (1.57 vs. 16.38\%) and 17.95\% OR-compliance. Scale does not equate to fairness, then: the 70B model amplifies rather than decreases skew. Generate-and-filter versions copy baseline behavior (e.g., GPT-4o 12.75 vs. 12.81\% OR-compliance), establishing that post-hoc filtering can't change an original skewed distribution—it only selects outputs already likely under the base model.
\subsection{Lexical Diversity}\label{sec:diversity}
\begin{figure}[t]
    \centering
    \includegraphics[width=\linewidth]{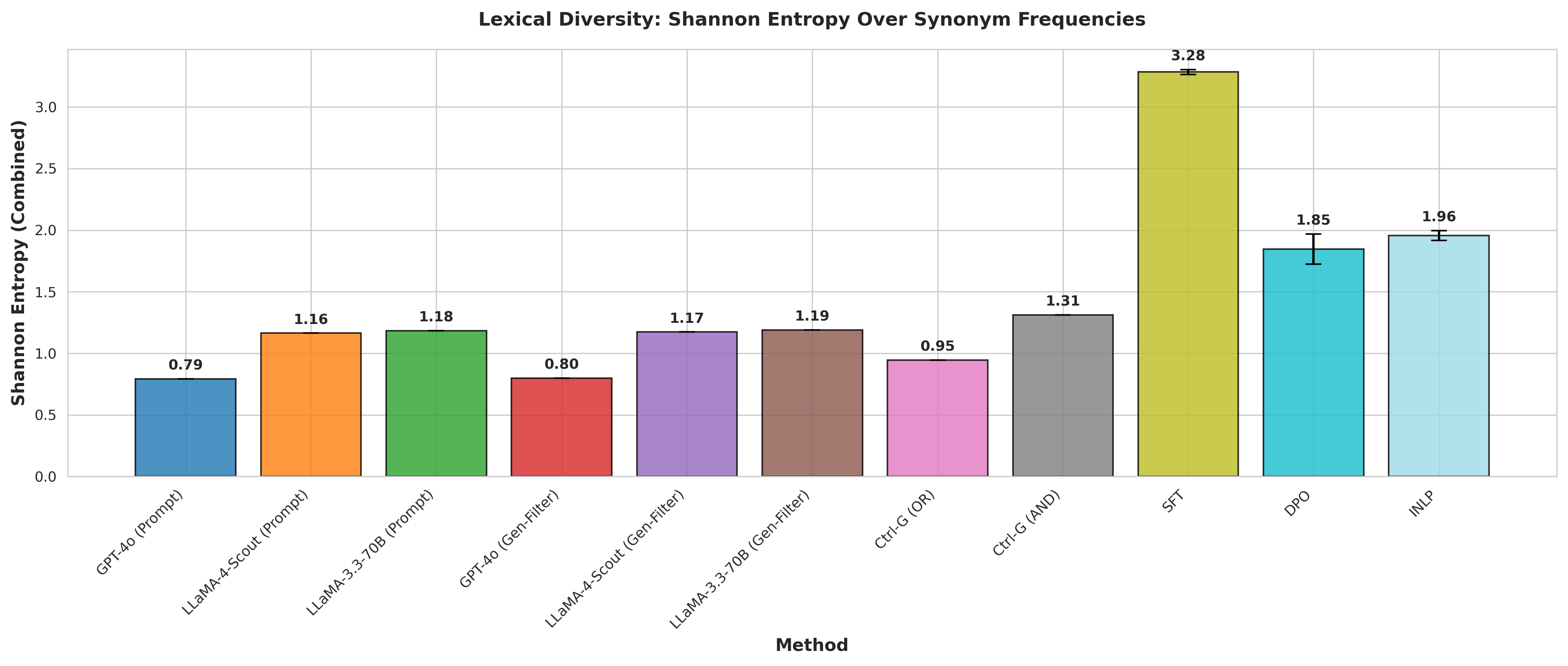}
    \caption{Diversity metrics across methods.}
    \label{fig:diversity}
\end{figure}

Figure~\ref{fig:diversity} presents Shannon entropy and distinct (agentic, communal) pair counts. Bare compliance is not sufficient; models vary significantly in obtaining it.\\ \textbf{Diversity vs. compliance trade-off.} SFT obtains near-optimal lexical diversity: total entropy 3.284 (agentic 3.298, communal 3.271) with 100 distinct pairs, the theoretical maximum. Entropy of order $\log_2(10) \approx 3.32$ indicates uniform sampling over vocabularies. SFT learns compositional structure rather than memorizing concrete templates. Ctrl-G (AND) gains only 13 unique pairs and combined entropy 1.313 (agentic 0.477, communal 2.150). Visual inspection shows duplicated wording (``confident and caring,'' ``confident and friendly,'' ``confident and supportive'') across professions. The low agentic entropy shows excessive use of a limited set of words (``confident,'' ``bold,'' ``assertive''). Strict symbolic control ensures coverage but discourages lexical exploration; the DFA offers many paths, but beam search favors high-probability clichés. DPO shows moderate lexical variety (entropy = 1.845, 24 distinct pairs) but almost no compliance. The apparent diversity stems from single-trait outputs; among the very few balanced completions, 24 unique pairs occur, indicating partial pattern learning without consistent prioritization of balanced traits. INLP gets 1.956 entropy and just 4 pairs, as to be expected from its 0.09\% compliance, hardly any residual traits left after projection. Baselines (prompt-only and filter) get 0.79--1.18 entropy and 0 pairs, producing traits independently but never in combination. Ctrl-G (OR) behaves identically (0.945 entropy, 0 pairs), confirming that disjunctive constraints do not induce composition.
\subsection{Fluency Analysis}
\begin{figure}[t]
    \centering
    \includegraphics[width=\linewidth]{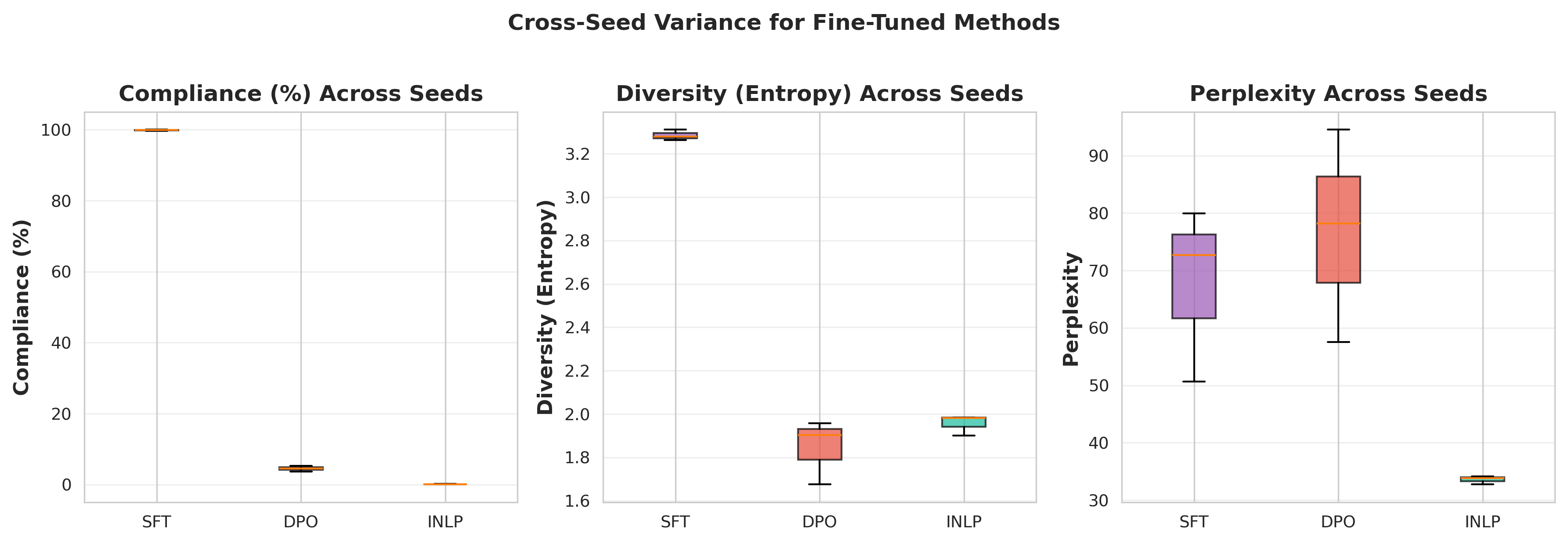}
    \caption{Cross-seed variance across methods.}
    \label{fig:fluency}
\end{figure}

\begin{figure}[t]
    \centering
    \includegraphics[width=\linewidth]{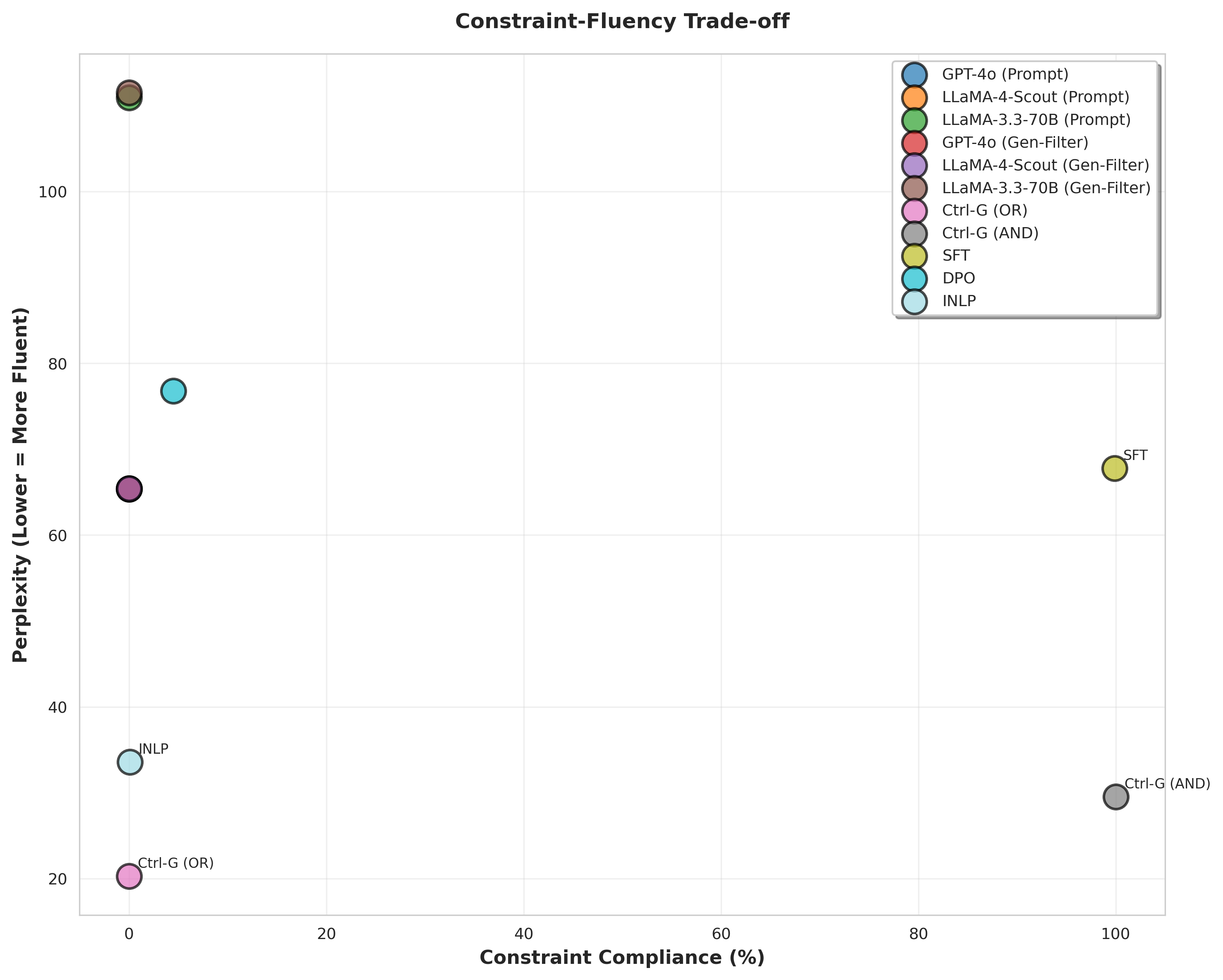}
    \caption{Trade-off between constraint strength and fluency.}
    \label{fig:frontier}
\end{figure}

Perplexity on GPT-2 Large is a fluency proxy. Figure~\ref{fig:fluency} illustrates the impact of the severity of constraint on naturalness.\\ \textbf{Fluency ranking.} Ctrl-G (OR) most fluent: mean perplexity 20.31 (median 19.12). Loosest OR constraint allows idiomatic single-trait completions (``The engineer was confident and driven''), which GPT-2 predicts high probability. Ctrl-G (AND) results in mid-fluency: 29.53 (mean) / 27.34 (median), a 45\% increase over OR with the same decoding. The improvement reflects the fluency penalty of conjunctive constraints: doubling up two descriptors within 8--15 words reduces predictability (``The architect was confident and supportive in their leadership''). Outputs remain, however, more fluent than most learning-based systems. INLP has strong fluency: 33.57 $\pm$ 0.60 perplexity—the lowest among fine-tuning methods. Because INLP modifies only a linear subspace, base-model distributions remain unchanged, producing natural but uninformative sentences with an abundance of neutral words. SFT has moderate fluency: 67.77 $\pm$ 12.43 perplexity (medians 45.32, 73.79, 68.41). Slightly above Ctrl-G or INLP but comparable to baseline models ($\approx 65$ for GPT-4o and LLaMA-4 Scout), its perplexity neither points to degradation nor overwhelmingly negative bias. SFT therefore preserves naturalness while achieving 99.87\% compliance, a good bias--fluency trade-off. The higher variance ($\sigma = 12.43$) indicates seed-level stylistic variation and not degradation. DPO does worst: 76.77 $\pm$ 15.14 perplexity (42 = 94.53, 123 = 78.23, 456 = 57.54). High variance reflects unreliable optimization; some runs generate fluent text, others generate degenerate outputs. Together with 4.51\% compliance, it indicates that preference learning cannot encode logical conjunctions without threatening fluency. Baseline models show unpatterned results: GPT-4o and LLaMA-4 Scout $\approx 65$ perplexity (fluent), whereas LLaMA-3.3 70B reaches 110.91 (according to GPT-2 reference). The larger model generates more sentences that are longer and less predictable, with stylistic deviation adding to more communal bias. As shown in Figure~\ref{fig:frontier}, these trends form a clear constraint-fluency frontier: models with stronger compositional control (SFT, Ctrl-G AND) incur higher perplexity, while those prioritizing naturalness (INLP, Ctrl-G OR) sacrifice compliance. This inverse relationship quantifies the fundamental tension between fairness and fluency observed across all control paradigms.
\subsection{Cross-Model Comparison}
Two trends follow from baseline analysis.\\ \textbf{1. Filtering does not fix bias.} Generate-and-filter models simulate prompt-only compliance (e.g., GPT-4o 12.75 vs. 12.81\%; LLaMA-4 Scout 20.55 vs. 20.33). Post-hoc selection can't redistribute traits because balanced outputs don't spontaneously appear.\\ \textbf{2. Scale does not imply balance.} The 70B LLaMA model (four times as large as the 17B variant) exhibits six times stronger communal bias (10:1 vs. 2:1) and $\approx$ 30\% reduced fluency (110.91 vs. 65.36 perplexity). Parameter count alone is not sufficient to counteract representational skew; scaling potentially doubles pre-training imbalances. Fairness-critical tasks should thus depend on empirical bias measurement instead of size as a proxy for neutrality.

\section{Discussion}
Our results demonstrate hidden distinctions in how control methods attain compositional constraints. Supervised fine-tuning attains near-optimal alignment (99.87\%) with the largest lexical diversity, but preference-based learning fails (4.51\%) with the same data and computation. These distinctions challenge the theory of equivalence among methods such as DPO or RLHF for alignment and demonstrate compositional constraints in preference-based training for control.
\subsection{Why Direct Preference Optimization Failed}
DPO's 4.51\% AND-compliance, in contrast to SFT's 99.87\%, reveals a conceptual incompatibility between preference learning and logical conjunctions.
\paragraph{Preference-learning bottleneck} DPO optimizes a binary classification loss over pairs of outputs, optimizing $\log\sigma\big(\beta[\log\pi_\theta(y^+\mid x)-\log\pi_\theta(y^-\mid x)]-\beta[\log\pi_\mathrm{ref}(y^+\mid x)-\log\pi_\mathrm{ref}(y^-\mid x)]\big)$, which specifies relative preferences (``balanced $>$ unbalanced'') but not absolute needs (``must have both traits''). A model can minimize this loss by slightly increasing the likelihood of balanced outputs while still generating mostly unbalanced ones, aligning with our observations. DPO achieves 33.71\% OR-compliance but only 4.51\% AND-compliance, indicating that it learns to produce individual trait words more frequently without combining them into balanced expressions.
\paragraph{Empirical evidence} Output distributions capture 18.36\% agentic-only and 10.85\% communal-only generations, while 66.29\% contain neither trait. The model evidently learned that each of the two types of traits is individually attractive but avoided mixing them and chose instead ``safe,'' neutral text. This aligns with previous research indicating that preference-optimized models have risk-averse policies that minimize rejection risk \citep{casper2023open}.
\paragraph{Explicit supervision is necessary} SFT, trained on 750 explicit positive instances, explicitly shows the model syntactic instantiations of balance (``confident and caring,'' ``ambitious yet empathetic''). The system thus learns the form of composition, not a preference ordering over examples. Compositional constraints (A AND B) cannot be learned from binary preferences (A $>$ $\neg$A); explicit positive examples are required. This limitation extends to all pairwise preference-based alignment methods, including RLHF, IPO, and KTO. These methods are impactful for subjective goals but are poor for discrete logical requirements. Future work should combine preference learning of subjective correctness with supervised cues for compositional correctness.
\subsection{Hard Constraints vs. Learning-Based Methods}
Comparing Ctrl-G, SFT, and INLP highlights trade-offs between control guarantees, efficiency, and generalization.
\paragraph{Ctrl-G (AND): Ideal compliance, bad diversity} The DFA guarantees 100\% AND-compliance, but at three costs: (1) Constraints must be enumerated in advance, limiting adaptability; (2) Lexical diversity shrinks to 13 unique (agentic, communal) pairs vs. 100 for SFT; (3) Small mis-specification causes drastic jumps, as exemplified by the OR variant (99.69\% OR, 0\% AND). Ctrl-G is best suited for rule-based generation tasks, such as regulatory or templated documents, where constraints are explicit and well-defined.
\paragraph{SFT: Pretty near ideal learning with generalization} SFT achieves 99.87\% conformity and ultimate diversity, just 0.13 points short of Ctrl-G, but with much more flexibility. With plenty of examples, the network acquires compositional reasoning and deduces novel combinations of synonyms and syntax (18 templates $\times$ 10 agentic $\times$ 10 communal $=$ 1,800 new forms from 750 training examples). This ability to generalize to new combinations makes SFT especially valuable for developing fairness criteria that cannot be exhaustively predefined.
\paragraph{INLP: Over-corrective but quick} INLP learns in $\approx$ 20s per seed (many hundreds of times more quickly than SFT) but only reaches 0.09\% compliance and removes $\approx$ 95\% of all trait words. Its linear-projection assumption wipes out entire semantic directions associated with gender, destroying both stereotypes and legitimate descriptions. INLP is applicable where neutrality is the goal (e.g., anonymized screening), but does not suffice for balanced representation tasks.
\subsection{The Diversity--Fluency Trade-off}
No strategy optimizes compliance, diversity, and fluency simultaneously; practitioners must select based on use case.
\paragraph{SFT: Balanced profile} SFT achieves 99.87\% compliance, entropy 3.284 (close to $\log_2 10 \approx 3.32$), and perplexity $\approx 68$, comparable to baseline fluency. It therefore suits applications where balanced, varied language is more valued than low perplexity: educational or clinical scenarios where representational equity is more desirable than stylistic polish.
\paragraph{Ctrl-G (AND): Compliance over variety} Ctrl-G has constraint satisfaction and relatively low perplexity ($\approx 30$), but only has 13 unique pairs. Beam search always selects high-likelihood patterns (``confident and caring''), producing bland text. This is at the expense of forgoing creative diversity for controlled contexts that need verifiable compliance.
\paragraph{Ctrl-G (OR): Fluency over composition} Disjunction enables the lowest perplexity ($\approx 20$) by single-trait, generic phrases. But with 0\% AND-compliance, it defaults on any compositional fairness goal, illustrating how best fluency typically comes at the expense of logical balance.
\paragraph{No free lunch} Tighter constraints inevitably reduce fluency by limiting high-probability continuations, while looser constraints enhance naturalness at the cost of increased bias. In practice, SFT or Ctrl-G (AND) offers the necessary balance for fairness-sensitive text generation, whereas weaker methods may suffice for exploratory or creative applications.
\subsection{Limitations}
Experiments focus on occupational titles: a well-studied source of gender stereotypes \citep{gaucher2011evidence}, but not the sole one. Follow-up research must examine other domains such as health care, education, or policy, in which equity involves other compositional dimensions (e.g., ``competent and empathetic''). Additionally, all tests use English models and Western taxonomies of characteristics. Cross-linguistic replication is necessary to determine if these failures generalize to languages with grammatical gender or other cultural pairings of agency and communion. We also only considered three instruction-tuned baselines (GPT-4o, LLaMA-4 Scout, LLaMA-3.3 70B) and one fine-tuned backbone (LLaMA-3.1 8B). Results may vary across architectures or training pipelines. However, the qualitative conclusion likely generalizes: preference learning does not capture logical composition, whereas supervision does. This follows from the design of their objective functions rather than from model-specific factors.

\section{Conclusion}
The paper demonstrates that compositional fairness in large language models is unachievable by preference-based alignment alone. Direct Preference Optimization and related methods, though strong for subjective alignment, impose only relative preferences and are not capable of internalizing logical conjunctions needed to produce balanced, multi-attribute text. Supervised fine-tuning, on the other hand, with noisily labeled explicit positive instances, has near-optimal compositional conformity with good lexical diversity and reasonable fluency, and can match hard constraints like Ctrl-G while generalizing and fitting in. These findings emphasize that control in LLMs driven by no preference but fairness is not an optimization problem but a representational one: logical structure enforceability requires supervision, not preference.

\section*{Ethical Considerations}
This work analyzes gendered language patterns in large language models using synthetic occupational prompts. No human subjects or sensitive personal data were used. Occupational terms and descriptors were selected from publicly available linguistic resources to study representational bias in aggregate, not to reflect real individuals. The study aims to improve fairness and transparency in language model behavior and does not attempt to alter or rank demographic groups.\\ Code and dataset are available at \href{https://github.com/atijmahesh/compositional-bias-control}{github.com/atijmahesh/compositional-bias-control}.

% Figures moved near first mention in text

\bibliographystyle{plainnat}
\nocite{*}
\bibliography{references}

\end{document}